\documentclass[nohyperref]{article}

\usepackage{microtype}
\usepackage{graphicx}
\usepackage{subfigure}
\usepackage{booktabs} % for professional tables

\usepackage{hyperref}

\usepackage[accepted]{icml2022}

\usepackage{amsmath}
\usepackage{amssymb}
\usepackage{mathtools}
\usepackage{amsthm}

\usepackage[capitalize,noabbrev]{cleveref}

\theoremstyle{plain}
\newtheorem{theorem}{Theorem}[section]

\theoremstyle{definition}
\newtheorem{definition}[theorem]{Definition}

\theoremstyle{remark}

\usepackage{breqn}
\usepackage{graphicx}
\usepackage{indentfirst}
\usepackage{amsmath,amsthm}
\usepackage{amsmath}
\DeclareMathOperator*{\argmax}{arg\,max}

\usepackage{pdflscape}
\hypersetup{hidelinks}
\usepackage{adjustbox}
\usepackage{sidecap}
\usepackage[utf8]{inputenc} % allow utf-8 input
\usepackage[T1]{fontenc}    % use 8-bit T1 fonts
\usepackage{hyperref}       % hyperlinks
\usepackage{url}    
\usepackage{enumitem}
\usepackage{booktabs}       % professional-quality tables
\usepackage{amsfonts}       % blackboard math symbols
\usepackage{amsmath}       % blackboard math symbols
\usepackage{nicefrac}       % compact symbols for 1/2, etc.
\usepackage{microtype}      % microtypography
\usepackage{lipsum}
\usepackage{graphicx}
\usepackage{subfigure}
\usepackage[font=small]{caption}
\usepackage{multicol}
\usepackage{multirow}
\usepackage{afterpage}
\usepackage{xcolor}

\usepackage[textsize=tiny]{todonotes}

\icmltitlerunning{RL4S : Reinforcement Learning For Survival}

\begin{document}

\twocolumn[
\icmltitle{Reinforcement Learning For Survival: A Clinically Motivated Method For Critically Ill Patients}



\icmlsetsymbol{equal}{*}

\begin{icmlauthorlist}
\icmlauthor{Thesath Nanayakkara}{tcn}

% \icmlaffiliation{tcn}{Department of Mathematics, University of Pittsburgh, Pittsburgh, United States of America}
% \icmlauthor{Firstname2 Lastname2}{equal,yyy,comp}
% \icmlauthor{Firstname3 Lastname3}{comp}
\end{icmlauthorlist}

\icmlaffiliation{tcn}{Department of Mathematics, University of Pittsburgh, Pittsburgh, United States of America}
\icmlcorrespondingauthor{Thesath Nanayakkara}{tcn10@pitt.edu}

\vskip 0.3in
]



\printAffiliationsAndNotice{}  % leave blank if no need to mention equal contribution

\begin{abstract}
There has been considerable interest in leveraging RL and stochastic control methods to learn optimal treatment strategies for critically ill patients, directly from observational data. However, there is significant ambiguity on the control objective and on the best reward choice for the standard RL objective. In this work, we propose a clinically motivated control objective for critically ill patients, for which the \emph{value functions} have a simple medical interpretation. Further, we present theoretical results and adapt our method to a practical Deep RL algorithm, which can be used alongside any value based Deep RL method. We experiment on a large sepsis cohort and show that our method produces results consistent with clinical knowledge.
\end{abstract}

\section{Introduction}
Recently, there has been an increased volume of research which try to learn optimal treatment strategies for critically ill and in particular for septic patients \cite{komorowski2018artificial,chen2019new,raghu2017deep,li2019optimizing,peng2018improving,festor2021enabling,nanayakkara2022unifying}, using Reinforcement Learning (RL) methods. Given the enormous mortality, morbidity and economic burden \cite{liu2014hospital,rhee2017incidence,paoli_reynolds_sinha_gitlin_crouser_2018}, the ambiguity regarding optimal treatment strategies and lack of accepted guidelines for treatment \cite{marik2015demise,jarczak2021sepsis}, such attempts are certainly justified. 

In this work, we will focus on applications where reduced mortality is the primary clinical goal. For such problems, there has been debate on optimal reward choices for the RL formulation. Indeed, some work have used exclusively terminal rewards (for example, $+/- 1$ depending on death or release or just a negative reward for death) \cite{komorowski2018artificial,li2019optimizing,killian2020empirical}, whilst others have used clinically motivated intermediate rewards \cite{raghu2017deep,peng2018improving,nanayakkara2022unifying}. Whilst just using terminal rewards does make sense as a clinical objective, such sparse reward choices induce high sample complexity, and all RL applications to medicine are performed in an \emph{offline} manner, using a fixed dataset of observed trajectories. In particular, for complex syndromes such as sepsis, given the enormous heterogeneity and complexities amongst patient trajectories, it is very unlikely that the extent and the variety of the currently available data will cover the feasible range of physiologic states in any case. Further, it is well known that even survivors face a significant readmission risk and a reduced life expectancy \cite{cuthbertson2013mortality,gritte2021septic}. Therefore, not all survivors are the same, and we may have to consider the physiologic health or even a physiologic expected life time of the survivors when they are released.

The current intermediate reward choices are mostly adhoc, and typically it is not verified whether maximizing cumulative discounted rewards is a reasonable clinical goal. Further, there is enough evidence in RL where reasonable looking reward choices have caused undesirable or even dangerous behavior \cite{amodei2016concrete,everitt2021reward}. In either case, the use of discount factors (which is necessary for mathematical guarantees) makes the interpretation of value functions opaque.

Thus, we propose a simple clinically motivated control objective for this problem: Maximizing the probability of surviving the ICU stay. We show how this objective could then be interpreted as a Q learning based RL problem with patient state, and action specific discount terms. Thereby, allowing us to use any Deep Q learning based algorithm with a one line modification. The \emph{Survival Q functions} also has a simple interpretation which can help in improving the trustworthiness of an RL agent, and provide \emph{some} explainability of recommended actions.

Further, the same theoretical properties as Q learning hold under mild assumptions. We, then experiment with this method using a large sepsis cohort and show qualitative differences between values and policies, compared with standard RL methods. We show that the scaled values are in particular, more consistent with clinical knowledge under our method.

In summary, in this article:
\begin{itemize}
    
    \item We introduce a new, survival focused objective for critically-ill patients. 
    
    \item We present theoretical results and then adapt this objective to a practical Deep RL algorithm.
    
    \item We experiment using a large sepsis cohort, and present how values are more consistent with clinical intuition under our scheme.
    
\end{itemize}
% \label{submission}

\subsection{Related Work}
As mentioned previously, there are a large volume of research which attempt to use RL or control for critical care applications \cite{liu2020reinforcement,komorowski2018artificial,raghu2017deep, yu2019reinforcement}.

However, for the best of our knowledge there is limited prior work which explore alternate control objectives \footnote{There have been risk sensitive RL methods, which optimize a different functional of the return rather than the expected value. However, these methods are also subject to a proper definition of rewards.} or  systematic criteria for defining RL rewards. \cite{prasad2020defining} define a class of reward functions for which high-confidence policy improvement is possible. The authors, identity a space of reward functions that yield policies that are
consistent in performance with the observed data. \cite{nanayakkara2022deep} learns a mortality risk score using semi supervised contrastive learning, and then use their risk score to define intermediate rewards as the decrease in risk between successive time steps.

Arguably, the closest to our work is Q learning approaches for censored data such as \cite{goldberg2012q}. However, their problem is fundamentally different to ours. They consider censored data, and define an objective which maximizes the survival time, taking the possible censoring into account. However, they focus on longer term problems and in contrast we focus on the shorter term, acute illnesses. We also have access to the end state of the patients, thus censoring isn't a major issue here.

Outside of medicine, \cite{ye2017survival} proposed a method, which aims to optimize the cumulative rewards in a constrained MDP, with a negative avoidance constraint. Their method uses a Negative Avoidance Function (NAF), which plays a role similar to a hazard function. However, apart from the higher level goal of prioritizing survival, the method proposed here is significantly different.

\section{Background}
We will start by briefly discussing the familiar RL framework and the additive control objective.
RL can be formalized by a Markov Decision Process (MDP) framework. This include state and action spaces $\mathcal{S},\mathcal{A}$, a (typically unknown) Markov probability kernel $p( |s,a)$, which gives the dynamics of the next state, given the current state and the action and a reward process with a kernel $r(|s,a)$.

Given a discount factor $\gamma$, the return is defined as the cumulative discounted rewards : $\sum_{t=1}^{\infty} \gamma^t r_t$, which is a random variable. In RL, the agent's performance is measured in terms of the return, and most of the attention has been focused on the \emph{expected} return.

Therefore, the \emph{value} of a policy $\pi$ ($V^\pi(s)$) at state $s$ is defined as the expected future rewards starting from state $s$, and following the policy $\pi$. 
That is :
\begin{equation}
    V^{\pi}(s)=\mathbb{E}_{p,\pi}[\Sigma_{t}\gamma^{t}r_{t}|s_{0}=s,\pi], \hspace{10pt} \forall s\in \mathcal{S}
\end{equation}

The Bellman equation for the value function can be written as:
\begin{equation}
 V^{\pi}(s)=\mathbb{E}_{p,\pi}[r+\gamma V^{\pi}(s')],
 \label{eq:bell_value}
\end{equation}

If $V^*$ is the optimal value function, $V^*$ satisfies the following Bellman optimality equation:
\begin{equation}
 V^{*}(s)=\sup_{\pi \in \Pi}\{\mathbb{E}_{p,\pi}[r+\gamma V^{*}(s')]\}
 \label{eq:bell2_value}
\end{equation}

Similarly, the \emph{state action value function} or $Q$ function can be defined as
\begin{equation}
    Q^{\pi}(s,a)=\mathbb{E}_{p,\pi}[\Sigma_{t}\gamma^{t}r_{t}|s_{0}=s,\pi,a_{0}=a],  \hspace{10pt} \forall s\in \mathcal{S}, a\in \mathcal{A}
\end{equation}
The $Q$ function can be interpreted as the expected return of starting at state $s$, taking the action $a$ and then following the policy $\pi$.

The following can then be verified.

The Bellman equation for the $Q$ function:
\begin{equation}
 Q^{\pi}(s,a)=\mathbb{E}_{p}[r]+\gamma \mathbb{E}_{p,\pi}[Q^{\pi}(s',a')],
 \label{eq:bell1_2}
\end{equation}
% \footnote{Note that here, $r$ is random variable distributed according to an appropriate conditional distribution, conditioned on both $s$ and $a$. Whilst  $r$ in \ref{eq:bell_value} is only conditioned on the state and the policy (If the policy is deterministic the interpretation becomes similar). However we don't make the conditioning explicit in our notation for simplicity.}

and the Bellman optimality equation for the $Q$ function:

\begin{equation}
 Q^{*}(s,a)=\mathbb{E}_{p}[r]+\gamma \mathbb{E}_{p}[\max_{a'\in A}Q^{*}(s',a')]
 \label{eq:bell_2_2}
\end{equation} (where $Q^{*}(s,a)$ is the optimal $Q$ function, and $s'$ denotes the random next state) 

Here, we have also implicitly assumed that the maximum exists for some $a \in \mathcal{A}$. If it doesn't, one can replace $\max$ with $\sup$.

Indeed, it can be shown that under some regularity conditions all four Bellman operators are contractions in $L^{\infty}$. So an iterative algorithm would converge to either the optimal value function or the policy induced value function.

\section{Reinforcement Learning for Survival}
\textbf{An idealized objective for critically ill patients}

Now, we will present an idealized, clinically motivated control objective for critically ill patients. We will start by defining the objective without any consideration of its usefulness as a computational method, and then refine it so that it can be adapted to a RL algorithm, with convergence guarantees.

We will assume the knowledge of a true discrete time conditional hazard (or survival) process. That is: suppose a patient's death is a (Markov) stochastic process, based on the patient state, and a given action. Thus, for each patient state, at each time $t$ there is a probability (discrete hazard) $h_{t}(s_t,a_t)=p(D_{t+1}=1|s_t,a_t,D_{t}=0)$ (where $D_{t}$=1 if the patient is dead at the end of the $t$ th time step and $0$ otherwise) of the patient dying within the next time step. We will further assume the hazard process is independent of the time $t$. Thus, we drop the subscript $t$ from $h_t(s,a)$ from now on, assuming the hazard process is stationary, but of course state and action dependent.

Now, for a given policy $\pi$, it is straightforward to compute the expected probability of a patient surviving their ICU stay as :
\begin{equation}
   \mathbb{E}_{p,\pi}[\prod_{t=0}^{H_s}(1-h(s_t,a_t))|\pi]
   \label{eq: surv_prob}
\end{equation}

Where, the expectation is taken with respect to the environment dynamics and the policy, the actions are $a_t \sim \pi(s_t)$ and $H_s$ is a state dependent random time, representing the remaining time at the ICU. Then, our control objective can be written as:

\begin{equation}
   \text{Maximize}, \hspace{5pt}
   \mathbb{E}_{p,\pi}[\prod_{t=0}^{H_s}(1-h(s_t,a_t))] \hspace{5pt} \text{such that} \hspace{5pt} \pi \in \Pi
   \label{eq: surv_obj}
\end{equation}

Where $\Pi$, is the class of policies considered.

Notice that the functional \ref{eq: surv_prob} is multiplicative, but we will not be using it in the same form any further. However, we note that traditional stochastic control literature have discussed multiplicative cost functionals \cite{bertsekas1996stochastic}. That work discusses DP-like algorithms and guarantees of optimal policies which hold for our survival objective Equation \ref{eq: surv_prob} (under known dynamics and an uniform finite horizon). However, we will take a different approach motivated by $Q$ functions. 

Let's define the \emph{survival Q function} : $Q_S^{\pi}(s,a)$ to be the probability of a patient with state $s$ will survive their ICU stay, given that the first action is $a$, and the policy $\pi$ is continued afterwards.

% That is :
\begin{definition}

\begin{equation}
     Q_S^{\pi}(s,a):=\mathbb{E}_{p,\pi}[\prod_{t=0}^{H_s}(1-h(s_t,a_t))|\pi,s_o=s,a_o=a]
\end{equation}

  \label{def:Q_S}
\end{definition}

Now analogous to Equation \ref{eq:bell_2_2}, we define the optimal survival Q functions as  $Q^{*}_S(s,a)$:
\begin{definition}

\label{def:Q_S_*}
\begin{equation}
    Q^{*}_S(s,a):=\sup_{\pi \in \pi}Q^{\pi}_S(s,a)
\end{equation}
  
\end{definition}

Now conditioning on the the event at $t=0$, the following two results follow immediately:

\begin{equation}
    Q^{\pi}_S(s,a)=(1-h(s,a))\mathbb{E}_{p,\pi}[Q_S^{\pi}(s',a')]
    \label{eq:Q_1}
\end{equation}

\begin{equation}
Q^{*}_S(s,a)=(1-h(s,a))\mathbb{E}_p[\max_{a'\in \mathcal{A}}Q_S^{*}(s',a')]
\label{eq:Q_*}
\end{equation}

With, for all $a \in \mathcal{A}$:

$Q^{\pi}_S(s,a),Q^*_S(s,a)=1$, when $s$ is a release state and,

$Q^{\pi}_S(s,a),Q^*_S(s,a)=0$ when $s$ is a death state.

% Note that \ref{eq:Q_*} can also be derived by conditioning on the outcome at $t=0$.
Now, let $R$ be an indicator variable such that $R(s)=1$ if $s$ is a release state, and 0 otherwise. We can interpret $R$ as a known, deterministic binary  function from $\mathcal{S} \to \{0,1\}$. 
% \footnote{Alternatively, we can think of $R$ as a known property or an annotation of the
%  state.}
Then, $Q^{*}$ satisfies the following relationship :
    \begin{equation}
    \label{eq:Q*_op}
        Q^{*}_S(s,a)=\mathbb{I}_{\{R(s)=1\}}+\mathbb{I}_{\{R(s)=0\}}(1-h(s,a))\mathbb{E}_{p}[\max_{a'\in A}Q_S^{*}(s',a')]
    \end{equation}

Implicit in Equation \ref{eq:Q*_op} is that for death states $h(s,a)=1$, so we don't have to explicitly consider that case. Before we describe the Deep RL algorithm we will present some theoretical results. For this, let's denote $F$ to be the set of real valued functions from $\mathcal{S} \times \mathcal{A} $ to $\mathbb{R}$, and define the operators $T_{\pi},T : F \to F$ as :

\begin{equation}
    T_{\pi}(J)(s,a)=\mathbb{I}_{\{R(s)=1\}}+\mathbb{I}_{\{R(s)=0\}}(1-h(s,a))\mathbb{E}_{p,\pi}[J(s',a')] ,
    \label{eq:T_pi}
\end{equation}

\begin{equation}
      T(J)(s,a)=\mathbb{I}_{\{R(s)=1\}}+(1-h(s,a))\mathbb{I}_{\{R(s)=0\}}\mathbb{E}_{p}[\max_{a' \in \mathcal{A}}J(s',a')],
    \label{eq: T}
\end{equation}

\begin{theorem}
\label{thm:bigtheorem}
Assume, the conditional hazard (for non release states) is uniformly bounded below by a positive number. Then, the operators $T_{\pi}$ and $T$ are contractions in the Banach space $B$ of bounded functions of $F$, under the sup-norm. Thus, they have unique fixed points.
\end{theorem}

The proof of Theorem \ref{thm:bigtheorem} follows with the exact same reasoning as results for analogous Bellman Q operators. However, we provide a proof in the Appendix A.

The contraction property of the optimal Survival Q function allows us to develop an experienced based, stochastic, \emph{Survival Q learning} algorithm, akin to Q learning. This algorithm is guaranteed to converge under the same assumptions as Q learning. We relegate this theorem (Theorem \ref{thm:thm2}) to Appendix A, due to space constraints.

As we noted earlier, the Survival Q function has a more straightforward interpretation than the regular Q functions (especially with intermediate rewards). That is : at each state $s$, and potential action $a$, $Q^{*}_S(s,a)$ represents the probability that the patient will survive their ICU stay, given that the action $a$ is taken at this time step and actions are taken optimally afterwards. Therefore, the agent has some capacity to explain the reasoning of each decision it recommends. However, we note that the quality of the interpretation depends heavily on the quality of the function approximators, training data and the approximate hazard model. Still, we believe compared with the existing methods, this is one of the advantages of our method. We could also interpret our method as an uncertainty aware method, which penalizes \emph{unlikely} survival by discounting the release by the likelihood of the survival, thus considering a form of aleatoric uncertainty. We will follow this insight and continue the discussion and possible modifications under Discussions.

\textbf{Reinforcement Learning for Survival (RL4S)}

Now, we can notice that Equation \ref{eq:Q*_op} can be compared with Equation \ref{eq:bell_2_2}, with zero intermediate \emph{rewards}, deterministic terminal \emph{rewards} and a state action specific \emph{discount} factor. Since, we have the knowledge of the end outcome of terminal states, we can use this relationship exactly as DQN \cite{mnih2015human} type algorithms leverage Equation \ref{eq:bell_2_2}. More specifically, we aim to parametrize the optimal survival Q function ($Q_S^{*})$ using function approximation based on Equation \ref{eq:Q*_op}. At terminal states the function is regressed into 1 or 0, and for every other state the left hand side is regressed to the the right hand side of Equation \ref{def:Q_S_*}, with the same convergence tricks as DQN. \footnote{Note, that this depends on a known hazard function, but there are several methods to learn an approximate hazard function, we will describe our choice in the experiments section.}

This insight, allows us to leverage any value based Deep RL algorithm, with a \emph{reward} \footnote{In our algorithm this is an indicator variable indicating if a patient has been released at the point or not. however the formulation fits into usual RL algorithms by interpreting this as a reward} where a) a final reward of $1$ is applied if and only if a patient is realised and b) 0 at all other time points. Whilst when interpreted as a reward, this choice is still sparse, using state, action specific survival probabilities instead of a uniform discounting term encodes information about the patient's condition.

\section{Experiments}
Now, we will conduct several experiments to investigate the performance of RL4S and to empirically compare the policies and values with other RL formulations. We will focus on the problem of administering vasopressors, and fluids for septic patients. This problem is well suited for our objective and is a popular choice for RL approaches \cite{raghu2017deep,komorowski2018artificial,li2019optimizing,killian2020empirical,nanayakkara2022unifying}. However, we emphasise that our focus here is to investigate our method and thus our results are preliminary and doesn't include many necessary steps needed before it can used for practical clinical decision support. For example, we strongly believe that any application of computational methods for clinical decision support should include (especially epistemic) uncertainty quantification, however we don't explore such results here. In particular, we do not claim that the learned policies are superior to that of the clinicians or previous RL efforts.

\subsection{Data Sources \& Prepossessing}
For all our analysis we used the MIMIC-III \cite{mimiciii,mimiciiidata} database and the same patient cohort which was used by Nanayakkara et al \cite{nanayakkara2022unifying}, including the representation learning described in that work. The cohort consisted of 18472 different patients out of which 1828 were non-survivors. All of these patients were adults ($\ge 17$), who satisfied the Sepsis 3 criteria \cite{singer2016third}. The excluded patients included patients who died at the hospital, but after release from the ICU, and patients who had more than 25\% missing values (vitals and scores) after creating hourly trajectories. This cohort resulted in 2596604 hourly transitions. The state space was 41 dimensional.

All the features were standardized for all work, and the missing values were imputed using a last value carried forward scheme, as long as the missingness was less than 25\% after creating hourly trajectories. We used the 9 dimensional discrete action space used in \cite{nanayakkara2022unifying}.

\subsection{RL4S}

Since RL4S depends on a known hazard model, we first describe the approximate hazard model we used.
 
\textbf{Hazard Model}: We used a simple feed forward neural network (or multi-layer perceptron (MLP)) to estimate the conditional hazard. By definition, the conditional hazard is the probability of the event (in this case death) occurring within a time step, given that the event hasn't occurred previously. Therefore, using the Markov assumption, we frame this as a classification problem of predicting whether a patient would die within $t$ and $t+1$, given the state $s_t$ and the action $a_t$. To satisfy the iid assumption used in stochastic gradient descent, for each batch, we first sampled the patients and then randomly sampled a patient state of that patient. To combat the heavy imbalanced nature of the problem (only 0.07\% of states were death states), we sampled non-survivors more frequently, and for a non-survivor, the patient state was taken to be the terminal state with 50\% probability and a random state with 50\% probability. Our architecture had two separate bases for the state and action and then the two representations were combined and sent through another small MLP head.

We then adapted the existing Deep RL algorithms (We used the distributional C51 algorithm \cite{bellemare2017distributional}, but it is trivial to use any value based algorithm). The only change in implementation and design required is that instead of a uniform discount factor, we have to use a state and action specific survival probability, analogous to a discount factor (and defining a form of rewards as described previously).

In addition to RL4S, we also experimented with standard RL with terminal rewards of $+/- 1$ depending of release or death and no intermediate rewards.

All the methods were trained using \cite{bellemare2017distributional} for 7 epochs with the same hyper-parameters except the lower and upper limits of the approximating discrete distribution \footnote{These were taken to be 0 and 1 for Survival RL, -1.5 and 1.5 for terminal only RL}. However, all methods displayed variation amongst recommended policies across weights saved after each epoch. Therefore for the value and policy results we present in the next section, we first averaged the value distributions of neural networks trained  for 5,6 and 7 epochs.

\section*{Results}
We will now discuss some results of the previously discussed experiments. We will start by investigating the $Q$ values (Survival Q values for RL4S) of both methods.

\begin{figure}[h!]
   \centering
    \includegraphics[scale=0.15]{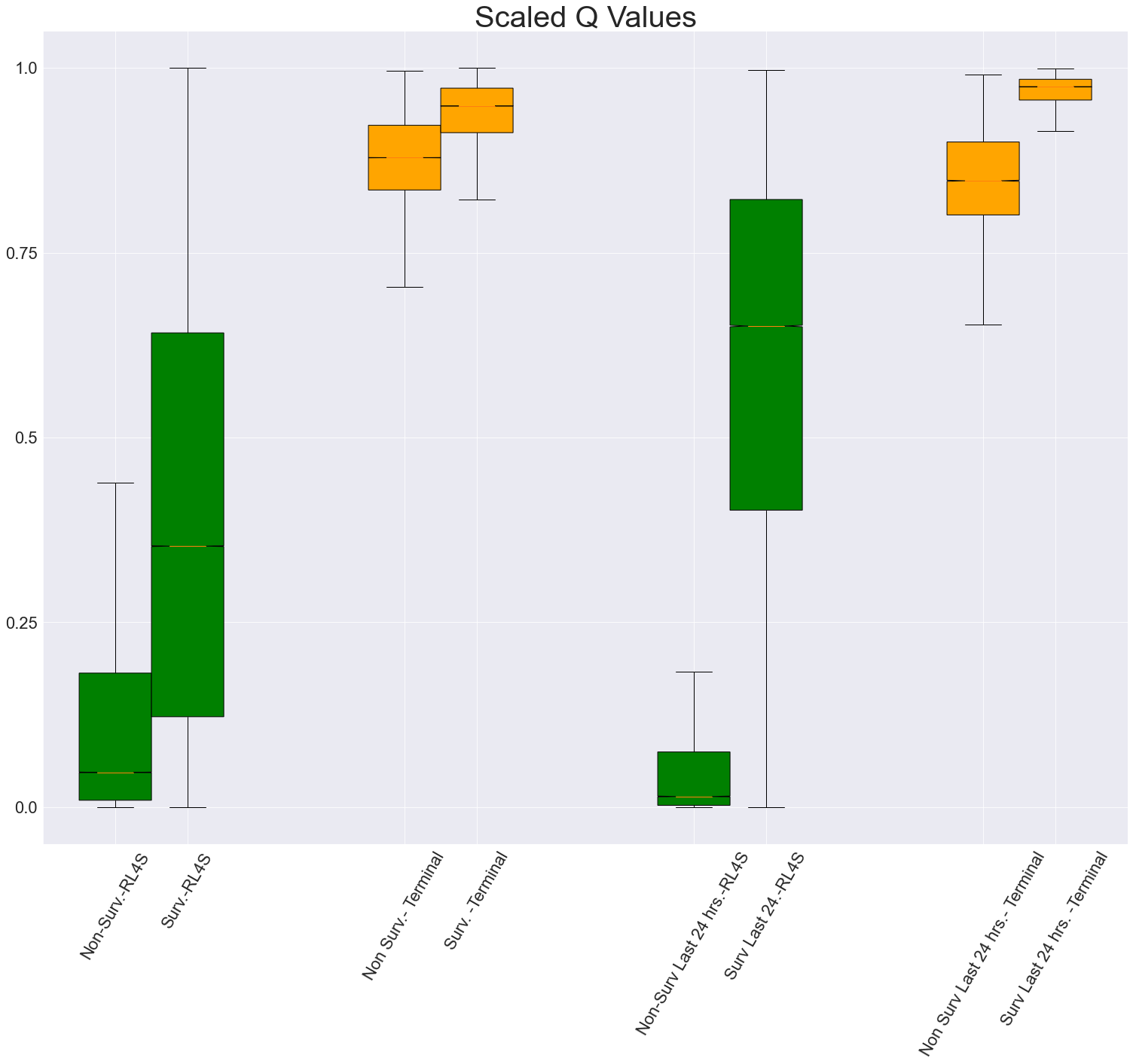}
    \caption{Box plots of Q values }
    \label{fig:box_surv}
\end{figure}

First, we consider the averaged $Q$ values (across actions and relevant states). Since the $Q$ values are defined and scaled differently in each case, we used a max-min scaling scheme -so the scaled $Q$ values are in between 0 and 1. We then stratified, these values by a) survivor, and non survivor states b) Last 24 hour states (before death or release) of each case. Figure \ref{fig:box_surv} presents these results using box plots. Here the green boxes denotes the Survival Q values of RL4S, the yellow : Q values for standard RL with terminal-only rewards. Intuitively, we expect the $Q$ values to capture the patient condition, and indicate the impending death or release at least when a patient is \emph{close} to each.

We can notice that there is a significant separation between survivor and non-survivor $Q$ values in RL4S. However, for RL with terminal-only rewards, even the median of the \emph{last 24 hr} non -survivor scaled $Q$ values is above 0.75. Considering the definition of the usual $Q$ function (For terminal rewards: Ignoring the discounting, the $Q$ value can be identified as a linear combination between expected release probability and expected death probability) this does not meet clinical intuition, as the models seem to be predicting survival even when the patients are close to death. In contrast RL4S in particular, seem to identify the higher mortality risk in advance.

We note that the main quantities of interest in RL algorithms, are not the values themselves but the difference between values of different actions. Therefore, it is possible for a method to overestimate $Q$ values, and yet correctly identify the correct ordering of $Q$ values (i.e. identify the optimal action order). However, explainability and trustworthiness are essential components of any automated medical decision making system. Value based algorithms attempt to learn optimal polices by estimating the values of states, and thus if the values themselves are inconsistent with clinical knowledge and observed outcomes, such a system is unlikely to be trusted. Therefore the results of RL4S seem to be more promising in this aspect. It is also important to note that our patient cohort was heavily dominated by survivors. A more balanced cohort \emph{could} result in more realistic $Q$ values. Another possibility is to bias the sampling scheme as explained in \cite{nanayakkara2022unifying}, by sampling death and near death states with higher probability.

% \begin{table}[h!]
% \caption{Actions (Act.) recommended by each method : RL with only terminal rewards (Term.), RL with intermediate rewards (Int.), RL4S and clinician's actions (Clin.)}
% \small
   
% %   \small
%     \centering
%     \begin{tabular}{||c|c|c|c|c|c||}
%      \hline \hline
%     Act. & Term. & Int. & RL4S & Clin.\\ \hline \hline
%     Flu 0 Vaso 0 & 64.9 & 80,5 & 81.4 & 27.8 \\ \hline
%     Flu 1 Vaso 0 & 11.2 & 16.3 & 11.1 & 55.5\\ \hline
%      Flu 0 Vaso 1 & 0.6 &  0.1 & 1.3 &  2.5\\ \hline
%       Flu 1 Vaso 1 &  23.3 & 3 & 6.1 & 14.1\\ \hline

%       \end{tabular}

%     \label{tab:tabl_4d_acts}
    
% \end{table}
Next, we will discuss selected interesting properties of recommended actions. Note that for each state $s$, we select the action $a$, which maximizes the $Q$ values. (i.e $a=\argmax_{a' \in \mathcal{A}} Q(s,a')$). We will present the full global action distribution in the appendices.

\begin{figure}[h!]
   \centering
    \includegraphics[scale=0.22]{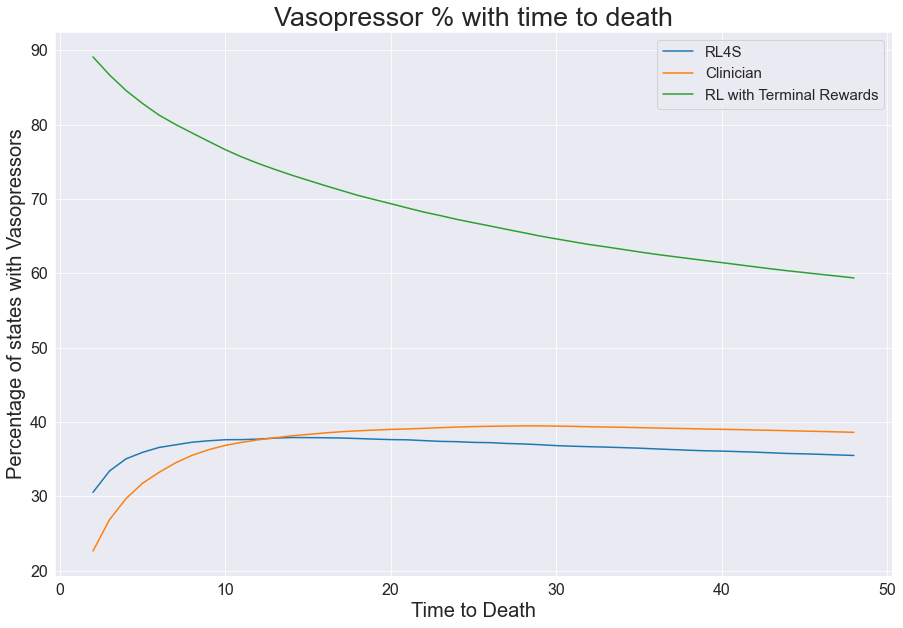}
    \caption{Percentage of states with vasopressors }
    \label{fig:vaso_per}
\end{figure}

A striking observation is illustrated in figure \ref{fig:vaso_per}. Here, we plot the percentages of states, with vasopressors recommended by each RL method, for non-survivors for different times to eventual death. Also, shown are the percentage of states for which the clinicians have used vasopressor therapy. The plots of RL4S and clinicians are remarkably similar, both even decrease as time to death decreases. However, for regular RL more vasopressors  are recommended as patients approach death, which is consistent with the results presented in \cite{nanayakkara2022unifying} for RL with intermediate rewards. They hypothesize that the decrease of states with vasopressors given by clinicians may be due to decisions that were made by the patient’s family to cease extraordinary measures. However, such information was not given to RL4S so it doesn't explain the behavior of RL4S. We plan to investigate the possible reasons in future work.

Unfortunately, evaluating policies in offline RL is an open problem with no satisfactory answers suited for critical care medicine \cite{gottesman2018evaluating}. Even, the current Off Policy Evaluation (OPE) methods are ill suited for intensive care medicine. Further, they are defined for a fixed reward choice making comparing  policies under two different objectives even more complicated. Thus, we don't make any claims that policies under one schemes is necessarily better at this point.

\section{Discussions \& Conclusions}
In this work, we introduced a control objective for RL applications in critical care medicine, which was motivated by the ambiguity of defining rewards. Indeed, the reward hypothesis is arguably the most fundamental component of RL and the only way to guide desired behavior of an agent. However, it is not immediate how rewards should be defined for most clinical decision making applications. Thus, we started from quantifying a reasonable clinical goal (i.e. maximizing the probability of survival) and developed a framework and an algorithm which can formalize this goal. We believe this objective is naturally suited to formalize the goal of reducing mortality. \footnote{Again, we emphasise that there are certainly other goals in critical care medicine, however we focus on problems where the primary goal is minimizing mortality risk. This certainly include a large class of problems}

One limitation of our method, is that it depends on an approximate hazard model. For our experiments, we used a simple MLP in a supervised learning setting to estimate the conditional hazard. Also evaluation of survival models are more complicated than standard supervised learning methods. However, given that survival analysis is a well researched area, there are several alternatives, including methods where medical knowledge can be encoded. There are also ways to reduce the effect of the learned hazard method. For example, one could define a hybrid method which considers survival of a short term horizon and then use a look-ahead value learned using standard RL methods. 

The similarity to Deep Q learning type algorithms, allows us to trivially implement a wide range of modifications and improvements to our method. For example, we can use most algorithms developed specifically for Offline RL. (For example, \cite{fujimoto2019off}) Informally, these methods attempt to learn policies which are sufficiently close to the behavioral distribution. Additionally, we can use Equation \ref{eq:Q_1} to define an Actor Critic method, instead of a pure value based method. Using distributional RL methods, we can naturally take environment uncertainty into account and modify Equation \ref{eq: surv_obj} by replacing the expectation operator by a risk sensitive measure (such as VAR or C-VAR) to define risk sensitive methods. In particular, methods designed for offline and risk sensitive problems such as \cite{Urpi21}, can be used.

Further, as we hinted earlier our objective has another interpretation which allows us to view it as an uncertainty aware method. To see this let's recall by Equation \ref{def:Q_S_*}, our objective can be seen as a standard RL objective, with rewards given if and only if a patient is released, and at each time, instead of using a fixed discounting term, the probability of survival $1-h(s,a)$ is used for discounting. Thus for each trajectory, the terminal reward is multiplied by the probability of surviving the ICU stay and thereby discounting unlikely releases more. This viewpoint allows us to investigate other avenues to incorporate Uncertainty Quantification, and possibly modify the objective.

Our initial experiments produced promising results. The Survival Q values seem to differentiate between survivor and non-survivor states and identify mortality risk in advance. However, as we have mentioned previously, comparing performance of different clinical RL methods using historical data is very challenging. Thus, further experiments and research have to be conducted before any stronger claims can be established. One possible way to evaluate the method would be to use a simulated environment of critically ill patients \footnote{Or a different environment with similar goals of survival}, and then compare the mortality rates under different methods, learned from a fixed set of trajectories. However, it is important to verify that any such environment will be sufficiently similar to the patient environment one is interested in, if not undesirable conclusions can follow. Thus, we defer these attempts to future work.

Finally, we note that stochastic control research has been historically dispersed amongst various mathematics, computer science, operations research and artificial intelligence communities. However, recently there has been an effort to unify these efforts in to a single framework \cite{powell2022reinforcement,meyn2022control}. We believe such an unified approach may result in methods specifically for healthcare and critical-care medicine.

\section*{Acknowledgements}
We are grateful for Professor Gilles Clermont (University of Pittsburgh School of Medicine, Department of Critical Care Medicine), Professor Christopher James Langmead (Carnegie Mellon University, School of Computer Science, Department of Computational Biology) and Professor David Swigon (University of Pittsburgh, Department of Mathematics) for their advice and the enlightening discussions which motivated this work.

\bibliography{example_paper}
\bibliographystyle{icml2022}

%%%%%%%%%%%%%%%%%%%%%%%%%%%%%%%%%%%%%%%%%%%%%%%%%%%%%%%%%%%%%%%%%%%%%%%%%%%%%%%
%%%%%%%%%%%%%%%%%%%%%%%%%%%%%%%%%%%%%%%%%%%%%%%%%%%%%%%%%%%%%%%%%%%%%%%%%%%%%%%
% APPENDIX
%%%%%%%%%%%%%%%%%%%%%%%%%%%%%%%%%%%%%%%%%%%%%%%%%%%%%%%%%%%%%%%%%%%%%%%%%%%%%%%
%%%%%%%%%%%%%%%%%%%%%%%%%%%%%%%%%%%%%%%%%%%%%%%%%%%%%%%%%%%%%%%%%%%%%%%%%%%%%%%
\newpage
\appendix
\onecolumn
\section{Proof of Fixed-Point Theorems}
\begin{proof}
First notice that in either case the image of $B$ is contained in $B$. i.e. $T,T_{\pi} :B \to B$.

We will first prove that $T_\pi, ($\ref{eq:T_pi}) is a contraction. 

For ease of notation we will introduce the following notation $\beta(s,a)= (1-h(s,a))$.
Then note that by assumption, there exist $\gamma$<1 such that $\beta(s,a)<\gamma , \hspace{5pt} \forall s,a$ with $R(s)=0$

Recall : $T_{\pi} : B \to B $ \hspace{5pt}
$T_{\pi}(s,a)=\mathbb{I}_{\{R(s)=1\}}+\mathbb{I}_{\{R(s)=0\}}(\beta(s,a))\mathbb{E}_{p,\pi}[J(s',a')]$

 Thus, for $J,J' \in B$
 
 $||T_\pi(J)-T_{\pi}(J')||_\infty$
 
 $=\sup_{s \in \mathcal{S}, a \in \mathcal{A}}|(T_\pi(J)(s,a)-T_\pi(J')(s,a)|$
 
%  $=\sup_{s \in \mathcal{S}, a \in \mathcal{A}}|((\beta(s,a))\mathbb{E}_{p,\pi}((J)(s,a)-(J')(s,a))|$
 
 $\le \sup_{s \in \mathcal{S}, a \in \mathcal{A}}|(\beta(s,a)) \mathbb{E}_{p,\pi}((J)(s,a)-(J')(s,a))|$
 
  $\le \gamma \sup_{s \in \mathcal{S}, a \in \mathcal{A}}|(J)(s,a)-(J')(s,a)|$
  
  $= \gamma ||J-J'||_\infty$

  The second part regarding the unique fixed point follows directly from the Banach contraction theorem, and the completeness of $B$.
  
  Now $T$ is defined as :
  
  $T(J)(s,a)=\mathbb{I}_{\{R(s)=1\}}+\mathbb{I}_{\{R(s)=0\}}(\beta(s,a))\mathbb{E}_{p}[\max_{a' \in \mathcal{A}}J(s',a')]$
  
  First notice that for any two functions $f_1,f_2 : \mathcal{X} \to \mathbb{R}$
  \begin{equation*}
      |\max_{x \in \mathcal{X}}f_1(x)-\max_{x \in \mathcal{X}}f_2(x)|\le \max_{x \in \mathcal{X}}|f_1(x)-f_2(x)|
  \end{equation*}
 
  Then, for $J,J' \in B$ and $s \in \mathcal{S}, a \in \mathcal{A}$
 
 $|T(J)(s,a)-T(J')(s,a)|$

 $|(\beta(s,a)\mathbb{E}_{p}[(\max_{a \in \mathcal{A}}(J)(s,a)]-\mathbb{E}_{p}[(\max_{a \in \mathcal{A}}(J')(s,a))]|$
 
 $=|\beta(s,a)\mathbb{E}_{p}[\max_{a \in \mathcal{A}}(J)(s,a)-\max_{a \in \mathcal{A}}(J')(s,a)]|$

 $\le (\beta(s,a))|\max_{a \in \mathcal{A}}(J)(s,a)-\max_{a \in \mathcal{A}}(J')(s,a))|$
 
 $\le (\beta(s,a))\max_{a \in \mathcal{A}}|(J)(s,a)-(J')(s,a))|$
 
  $\le ((\beta(s,a))\sup_{s \in \mathcal{S}, a \in \mathcal{A}}|(J)(s,a)-(J)(s,a)|$
  
  $< \gamma ||J-J'||_\infty$
  
 Now taking the supremum over $s \in \mathcal{S}, a \in \mathcal{A}$, we get that,
 $||T(J)-T(J')||_\infty \le \gamma ||J-J'||_\infty$
 
 Again, the fixed point property follows.

\end{proof}
\section{Stochastic Approximation Theorem}
\begin{theorem}
\label{thm:thm2}
If $(s_k,s'_{k},a_k,h_k(s_k,a_k),R_k) \hspace{5pt} k \in \mathbb{N}$ is a set of experience tuples, generated from the underlying patient distribution. Where $R$ is an indicator variable such that $R(s)=1$ if the patient is released at this state and 0 otherwise.

Suppose :

    $\alpha_k, \hspace{5pt} k \in \mathbb{N}$ is a sequence of positive real numbers satisfying the Robbins Monro conditions \cite{robbins1951stochastic}, (for state, action pairs $s_k,a_k$) :
    
    $\Sigma_{k=0}^{\infty}\mathbb{I}_{\{s=s_k,a=a_k\}} \alpha_k= \infty$ and $\Sigma_{k=0}^{\infty}\mathbb{I}_{\{s=s_k,a=a_k\}} \alpha_k^2< \infty.$ with probability 1 for all $s \in \mathcal{S}, a \in \mathcal{A}$.

Then, the algorithm defined by $Q^0_S(s,a)=0$ and: 

$Q^{k+1}(s,a)=(1-\alpha_k) Q_S^{k}(s,a)+(\alpha_k)\mathbb{I}_{\{s=s_k,a=a_k\}}[\mathbb{I}_{\{R(s)=1\}}+\mathbb{I}_{\{R(s)=0\}}\beta(s,a)\max_{a' \in \mathcal{A}}Q_S^{k}(s',a')]$

Converges to $Q_S^{*}(s,a)$ with probability 1.

\end{theorem}

The proof of the above theorem is also analogous to the corresponding convergence results of temporal difference methods and Q learning. However, a full proof, with the relevant background would be too lengthy for this text. We refer to \cite{borkar2009stochastic,bdr2022} for a general stochastic approximation results, and convergence proofs of Q Learning method \cite{watkins1992q}. 
% \newpage
\section{Implementation Details}

We used the standard C51 training algorithm as in \cite{bellemare2017distributional}. Q network was a multi-layer neural network.  We use a target network for all methods include RL4S, and update the target networks using polyak target updating with $\tau=0.005$. (i.e. after every iteration/training step we set the target network weights to a linear combination of it's own weights, weighted by (1-$\tau$) an the Q network weights, weighted by $\tau$). This kind of target network is common amongst all deep Q learning, algorithms.
We used the following hyper-parameters and optimization choices for the c-51 algorithm. As we mentioned previously, the maximum and minimum values of the approximating distribution and the discount factor for RL4S, were the only hyper-parameters which were not shared by all the methods.

\begin{table}[ht!]
\small
    \centering
    \caption{RL algorithm hyper-parameters}
    \begin{tabular}{||c | c|| }
        \toprule

        %  \\ %\multicolumn{3}{c}{Metric 1} & Metric 2\\
        % \addlinespace[3pt]
        % \cmidrule(lr){2-4} \cmidrule(lr){5-5} \\
        {Hyper-Parameter} &  Value \  \\ \hline
        \midrule
        Support size & 51 \\ \hline
        $\gamma$ & 0.999\\ \hline
        Batch size &  124 \\ \hline
        Number of iterations &  51932 \\ \hline
        Optimizer & Adam\\ \hline
        Learning rate & $3\times10^{-4}$\\ \hline
        $\tau$ & 0.005\\ \hline
        \bottomrule
    
    \end{tabular}
    
\end{table}

As mentioned previously, the hazard model was treated as a standard classification problem. All the optimizations were conducted using Adam \cite{kingma2014adam}.

For both the hazard model and RL the state consisted of :

\begin{itemize}
    \item \textbf{Demographics}: Age, Gender, Weight.
    \item \textbf{Vitals}: Heart Rate, Systolic Blood Pressure, Diastolic Blood Pressure, Mean Arterial Blood Pressure, Temperature, SpO2, Respiratory Rate.
    \item \textbf{Scores:} 24 hour based scores of, SOFA, Liver, Renal, CNS, Cardiovascular
    \item \textbf{Labs:} Anion Gap, Bicarbonate, Creatinine, Chloride, Glucose, Hematocrit, Hemoglobin, Platelet, Potassium, Sodium, BUN, WBC.
    \item \textbf{Latent States}: (see \cite{nanayakkara2022unifying}) Cardiovascular states and 10 dimensional lab history representation.

\end{itemize}
\newpage
\section{RL4S : Recommended Actions}

\begin{table}[h!]
\caption{Percentages of actions (Act.) recommended by RL4S and Clinicians}
\small
   
%   \small
    \centering
    \begin{tabular}{||c|c|c||}
     \hline \hline
    Act. & RL4S & Clinician\\ \hline \hline
    Flu 0 Vaso 0 & 59.89 & 27.78 \\ \hline
    Flu 1 Vaso 0 & 3.79 & 23.70\\ \hline
     Flu 2 Vaso 0 & 17.97 &  31.78\\ \hline
     Flu 0 Vaso 1 &  3.29 & 1.29 \\ \hline
     Flu 1 Vaso 1 &  2.98 & 3.28 \\ \hline
     Flu 2 Vaso 1 &  10.57 & 3.98 \\ \hline
     Flu 0 Vaso 2 &  0.55 & 1.26\\ \hline
     Flu 1 Vaso 2 &  0.91 & 2.51\\ \hline
     Flu 2 Vaso 2 &  0.01 & 4.40\\ \hline

      \end{tabular}

    \label{tab:tabl_acts}
    
\end{table}

%%%%%%%%%%%%%%%%%%%%%%%%%%%%%%%%%%%%%%%%%%%%%%%%%%%%%%%%%%%%%%%%%%%%%%%%%%%%%%%
%%%%%%%%%%%%%%%%%%%%%%%%%%%%%%%%%%%%%%%%%%%%%%%%%%%%%%%%%%%%%%%%%%%%%%%%%%%%%%%

\end{document}